\documentclass[conference]{IEEEtran}
\usepackage{cite}
\usepackage{amsmath,amssymb,amsfonts}
\usepackage{algorithmic}
\usepackage{graphicx}
\usepackage{textcomp}
\usepackage{xcolor}
\usepackage[hyphens]{url}
\usepackage{microtype} 

\def\BibTeX{{\rm B\kern-.05em{\sc i\kern-.025em b}\kern-.08em
    T\kern-.1667em\lower.7ex\hbox{E}\kern-.125emX}}

\title{PDR-CapsNet: an Energy-Efficient Parallel Approach to Dynamic Routing in Capsule Networks}

\author{\IEEEauthorblockN{ Samaneh Javadinia}
\IEEEauthorblockA{\textit{ECE Department}}
\textit{University of Victoria}\\
Victoria, BC, Canada \\
samanehjavadinia@uvic.ca
\and
\IEEEauthorblockN{ Amirali Baniasadi}
\IEEEauthorblockA{\textit{ECE Department}}
\textit{University of Victoria}\\
Victoria, BC, Canada \\
amiralib@uvic.ca}

\begin{document}
\maketitle
\thispagestyle{plain}
\pagestyle{plain}


\begin{abstract}
    

    Convolutional Neural Networks (CNNs) have produced state-of-the-art results for image classification tasks. However, they are limited in their ability to handle rotational and viewpoint variations due to information loss in max-pooling layers. Capsule Networks (CapsNets) employ a computationally-expensive iterative process referred to as dynamic routing to address these issues. CapsNets, however,  often fall short on complex datasets and require more computational resources than CNNs. To overcome these challenges, we introduce the Parallel Dynamic Routing CapsNet (PDR-CapsNet), a deeper and more energy-efficient alternative to CapsNet that offers superior performance, less energy consumption, and lower overfitting rates. By leveraging a parallelization strategy, PDR-CapsNet mitigates the computational complexity of CapsNet and increases throughput, efficiently using hardware resources. As a result, we achieve 83.55\% accuracy while requiring 87.26\% fewer parameters, 32.27\% and 47.40\% fewer MACs, and Flops, achieving 3x faster inference and 7.29J less energy consumption on a 2080Ti GPU with 11GB VRAM compared to CapsNet and for the CIFAR-10 dataset.

\end{abstract}


\section{Introduction}

    
    Capsule Network (CapsNet) performs computations at the capsule level.  A capsule is a small group of neurons and is used to detect a particular object inside an image. This detection is performed by producing an output vector. The length of this vector represents the estimated probability that the object is present. The vector's orientation encodes the object's pose parameters (e.g., precise position, rotation). These features make CapsNet more suitable for object detection and image segmentation applications. CapsNet relies on an iterative routing-by-agreement mechanism to achieve this. Sabour et al. \cite{sabour2017dynamic} used this routing method as a replacement for max-pooling. In contrast to max-pooling, dynamic routing does not lose information.

    \par
\begin{figure}[htp]
        \centering
        \includegraphics[keepaspectratio, scale=0.5] {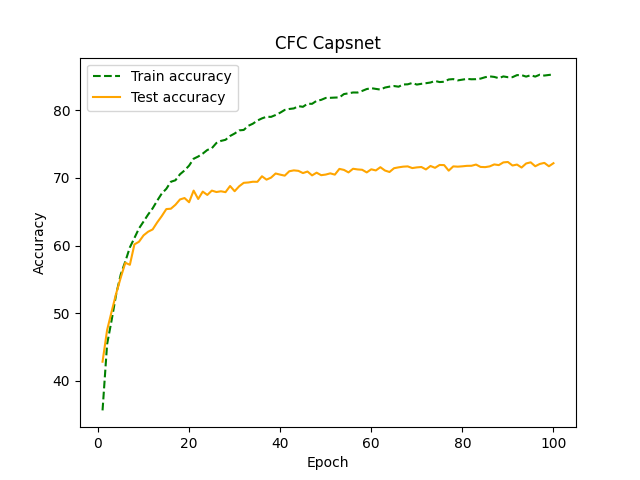}
        \caption{ CFC-CapsNet train and test accuracy for the CIFAR-10 dataset.}
        \label{fig:cfcconvergence}
    \end{figure}

 
     Convolutional Fully-Connected Capsule Network (CFC-CapsNet) is a CapsNet-based network. We build on this network and add to its efficiency. The CFC layer in this network uses all the channels corresponding to every pixel to create an associated capsule using the Convolutional Fully-Connected approach. One of the shortcomings of CFC-CapsNet is the shallowness of its architecture, resulting in underfitting and low accuracy. Figure \ref{fig:cfcconvergence} shows the corresponding accuracy for CFC-CapsNet. The loss of accuracy for the training set indicates underfitting. To address this drawback, we introduce parallel dynamic routing to deepen the network without facing overfitting. Moreover, we take an energy-aware approach and provide better energy efficiency compared to CFC-CapsNet.


    \par 
    Accuracy could sometimes be improved by using more data. Augmentation can do this when implementing a neural network for datasets such as the CIFAR-10. An alternative is to build a  deeper (and consequently more complex) network. This is achieved by scaling up the network's depth or width: increasing the number of layers or channels. This is usually practical when accuracy loss in test data stems from underfitting in train data\cite{szegedy2015going}. In PDR-CapsNet, we enhance accuracy by adding multiple paths with various feature extraction depths, including parallel dynamic routing components. Some of these paths are deeper than others providing the network with an opportunity to improve accuracy. Other paths are shorter and are employed to avoid excessive usage of deep paths which can lead to overfitting. In essence, we achieve high accuracy and energy efficiency by selecting the appropriate path intelligently. 
    Note that these multiple paths can enhance network performance by learning different aspects of images and retaining more valid information than the single path. At first, the added complexity boosts accuracy by reducing underfitting. However, when overfitting begins, the network chooses simple paths with lower parameter numbers as a prevention measure. 
    
    \par  The main difference among the multiple paths used in our approach is the number of feature extraction layers. We produce this choice by employing three alternative paths in terms of complexity throughout learning (more details in Section 4).

    \par These three alternatives are represented by the branches which feed three parallel dynamic routing paths, representing the image in various sizes. This improves generalization and brings about better equivariance.

    \par Moreover, we propose two methods to enhance CapsNet's computational structure. One way to improve computational efficiency is by replacing a 9*9 convolution with a 3*3 one.  
    The fewer parameters lead to lower computational complexity. In addition, 9*9 and 3*3 convolution layers decrease the dimension of the output image by eight and two neurons, respectively. 

\begin{figure}[htp]
        \centering
        \includegraphics[keepaspectratio,scale=0.15]{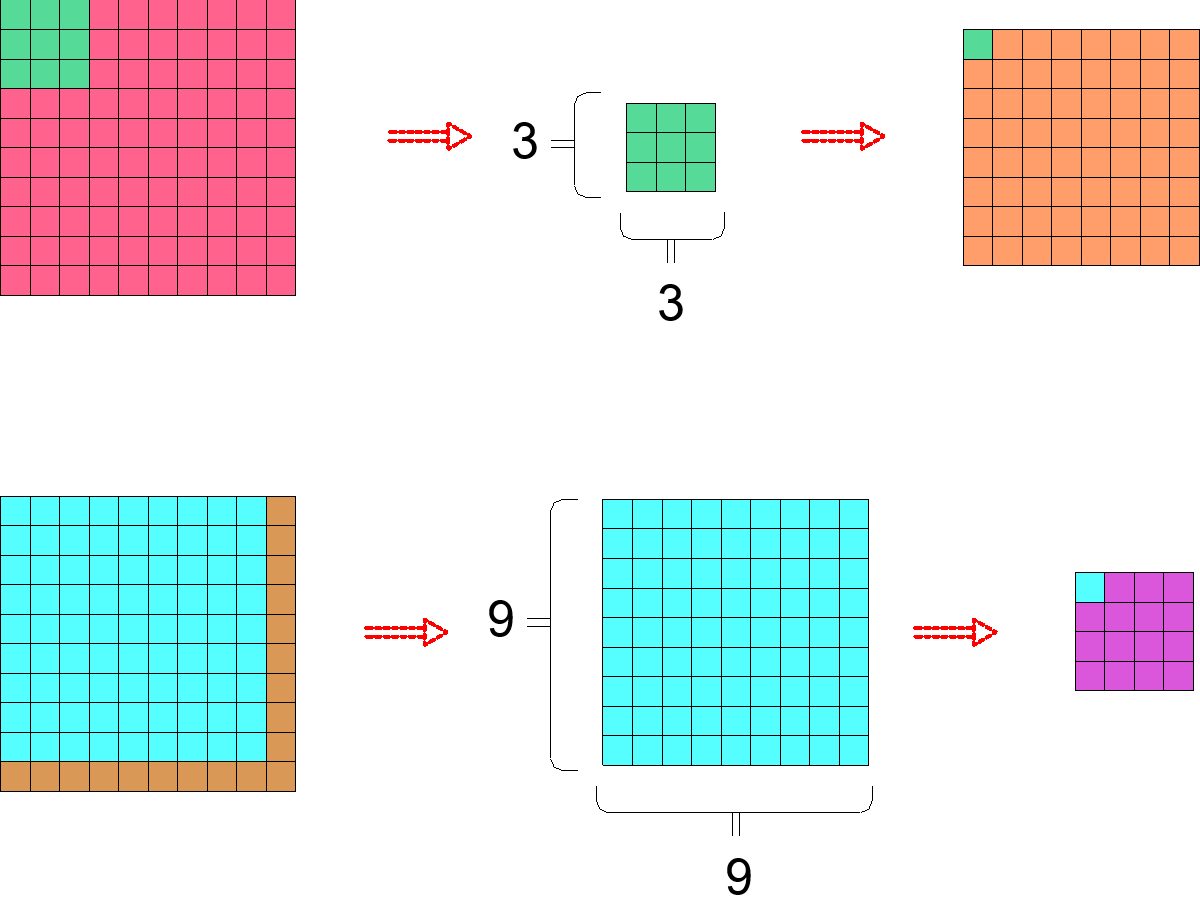}
        \caption{  Receptive field of 3*3 vs 9*9 convolution.The receptive field is important as it reflects the image information.
    }
    \label{fig: receptive}
    \end{figure}

    Employing 3*3 convolution layers also shrinks the region's size in the input that produces every neuron in the output (also referred to as the receptive field). In a 9*9 convolution layer, \( 9*9*n_c \) (\(n_c:\)the number of input channel) neurons participate in producing every output neuron. When the kernel size is 3*3, the number of participating neurons is \( 3*3*n_c \) (Figure \ref{fig: receptive}). The receptive field reduction can affect performance since input neurons outside the receptive field of an output unit do not affect the value of that unit.   In essence, we choose 3*3 convolution layers to gradually decrease the dimension of the image as its benefits outweigh the cost of the reduction in the receptive field. 

 \footnote {We use the following formula to describe the output size after a convolution and calculate the dimension reduction:  
    \[ ((W-f+2P)/S)+1 \]
     W: the input size\\
     f: the filter size\\
     P: the padding \\
     S: the stride\\
     Consider \( W = 32, P = 0 and S = 1\)\\
     The output size after a 3*3 convolution \( f=3\) is  \((((32-3+2*0)/1)+1)= 30 \) whose dimension is two less than the input. Nonetheless, the output size is \((((32-9+2*0)/1)+1)= 24\) after implementing a 9*9 convolution \( f=9\) with a dimension which is eight less than the input. This is why we should use a 3*3 convolution layer four times \( 2*4=8\) to reduce the dimension to the same as a 9*9 convolution layer. The formula for the number of parameters in a convolution layer is as follows:  
    \[ (f*f*n_c+1)*K\] 
    f: the filter size \\  
    \(n_c\): the number of input channel \\
    \(k\): the number of filters \\
    Consider \(n_c=k=256\)\\
    If we use four 3*3 convolution layers, the number of parameters will be: \\
    \( ((3*3*256+1)*256) * 4 \approxeq 2.36M\) \\However, a 9*9 convolution layer results in more parameters:\\
    \( (9*9*256+1)*256 \approxeq 5M\).}

    \par
\begin{figure}[htp]
        \centering
        \includegraphics[keepaspectratio, scale=0.15]
        {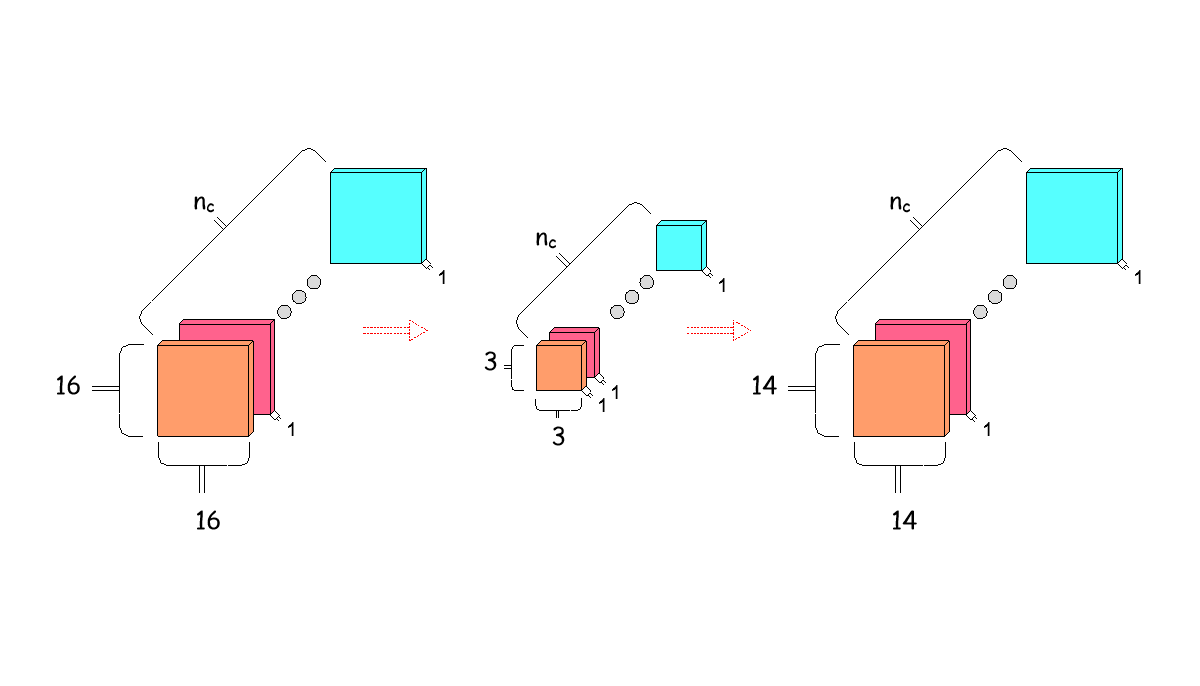}
        \caption{Depth-wise part of Depth separable convolution.
    Every filter slide over a single channel of input, creating a single channel of output with reduced dimension. Accordingly, the method learns the relation among pixels of a single channel of the input image. 
    }
    \label{fig: depth}
    \end{figure}

\begin{figure}[htp]
        \centering
        \includegraphics[keepaspectratio, scale=0.15]{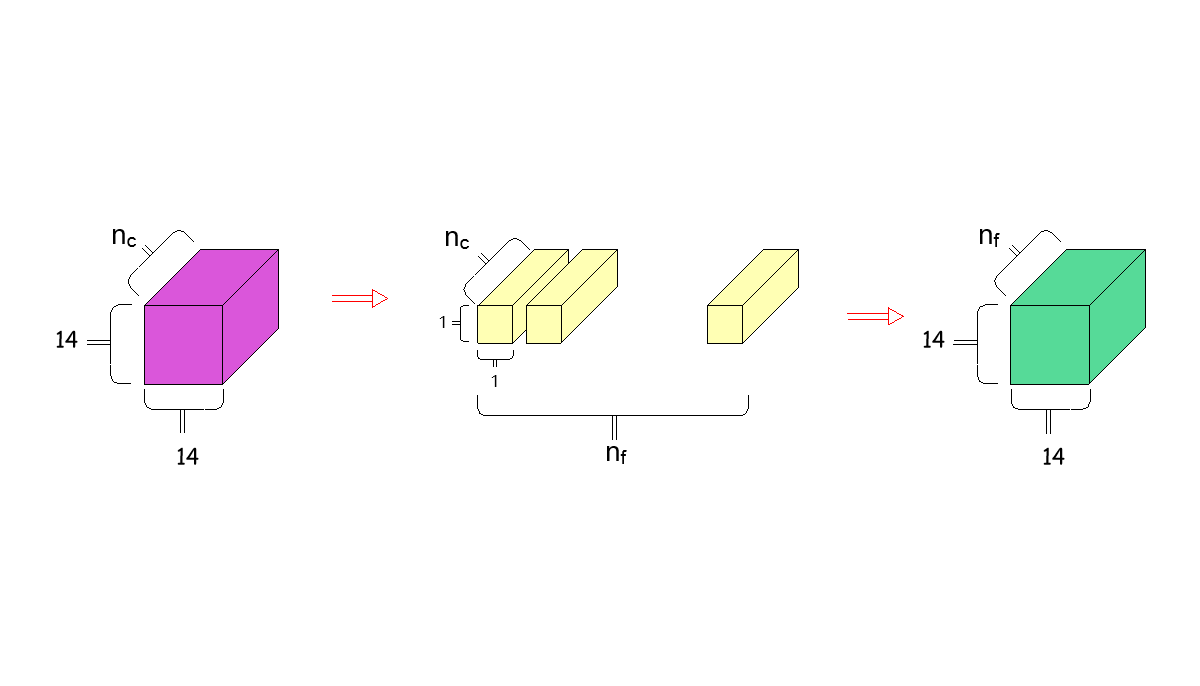}
        \caption{Point-wise part of Depth separable convolution.
    Every filter slides over the input pixel by pixel, creating pixels of output. Various filters result in various output channels. Accordingly, the point-wise convolution learns the relation among neurons of various channels related to every pixel.  }
        \label{fig: point}
    \end{figure}

 We also take advantage of reducing the number of parameters and computation complexity by using depth-wise separable convolution as shown in Figures \ref{fig: depth} and \ref{fig: point}. This convolution splits a kernel into two separate kernels performing two convolutions: the depth-wise convolution and the point-wise convolution. 
    Therefore, we reduce the image size and change its channel number using two separate convolutions. The depth-wise convolution also allocates different weights to different channels of the feature map, effectively highlighting more significant features.

    \par PDR-CapsNet's reduced computational complexity and shorter training and inference times contribute to its energy efficiency. By minimizing the number of operations required to execute the network, we reduce the energy consumed by the processor. Additionally, shorter training and inference times translate to less time spent processing, resulting in further energy savings. The energy efficiency of a neural network is closely related to its computational demands, with the number of multiply-accumulate operations (MACs) and floating-point operations (FLOPs) being common measures of these demands. These factors demonstrate and indirect measure of power reduction, while other factors such as memory access patterns and hardware utilization also play a role.
    \par One potential advantage of the proposed architecture is that it may offer improved scalability compared to traditional multi-GPU approaches in certain scenarios. By scalability, we refer to the ability of a system to handle increasing workloads without compromising performance. While it is true that adding more GPUs can increase compute power, it is also important to consider the overhead associated with inter-GPU communication and synchronization. This communication overhead can become a bottleneck in multi-GPU systems, limiting the overall scalability of the system. By contrast, a single GPU implementation avoids the overhead of inter-GPU communication and synchronization, allowing the GPU to focus solely on computation. Therefore the single GPU can potentially handle a larger workload without being limited by communication overhead, making it a more scalable solution in certain situations. Additionally, a single GPU implementation can be more energy-efficient compared to multi-GPU systems, which can be important for applications that require continuous operation.
       
    \par 
    In summary, we make the following contributions:
    
    \begin{itemize}
    \item 	We develop a highly accurate CapsNet-based network. Our network performs competitively as it takes advantage of the multiple branches concept introduced by  Szegedy et al.\cite{szegedy2015going}. These branches make the capsule network deeper while undertaking generalization tasks.
    
    \item 	We improve the conventional CapsNet in terms of the number of parameters and accuracy, Precision, Recall and F1-score. Our improvements lead to generating capsules representing parts of the image in various sizes which include different features of the image. We improve CapsNet's speed as we process fewer yet more powerful capsules. 
    
    \item 	We achieve parameter reduction using 3*3 convolutional layers and depth separable convolutions. We produce 87.26 \% fewer parameters, 32.27\% and 47.40\% fewer MACs and Flops and therefore less computational complexity compared to the conventional CapsNet. Moreover, we enhance accuracy from 71.69\% for CapsNet to 83.55\% for the CIFAR-10 dataset.

     \item 	We regularize the network and reduce the need for dropout, taking advantage of Batch Normalization \cite{ioffe2015batch}. We employ Batch Normalization (BN) right after every convolution and before the activation.


    \end{itemize}

    This paper is organized as follows:
    We discuss related works in Section II. We review the background in  Section III. We present Parallel Dynamic Routing CapsNet (PDR-CapsNet) in Section IV. We present experimental results in Section V. We conclude the paper in Section VI.

\section{Related Works}

    \par 
     Since the introduction of the original CapsNet, different variants have been suggested. Shiri and Baniasadi\cite{eshiri2021convolutional} propose Convolutional Fully-Connected Capsule Network (CFC-CapsNet). CFC-CapsNet takes advantage of a new CFC layer to create capsules. This layer produces fewer capsules compared to the conventional CapsNet. Employing these capsules results in higher accuracy compared to CapsNet. Better accuracy, faster training and a smaller number of parameters compared to the original CapsNet are other results of CFC-CapsNet.
    
    \par Shiri and Baniasadi \cite{shiri2021capsnet} propose LE-CapsNet. They introduce a new method of feature extraction. This network extracts features in different numbers of convolutional layers(also referred to as scales). This is done using a module referred to as Primary Capsule Generation (PCG) which extracts features from the input image. This module provides low-level and high-level capsules which effectively represent the input image. They also use CFC layers, an enhanced decoder in this network to improve performance further. This network achieves higher accuracy, faster training and interference and lower parameter number in comparison to CapsNet.

    \par Rosario et al.\cite{do2019multi} propose Multi-Lane CapsNet (MLCN). This network proposes Independent parallel lanes which create different Primary Capsules responsible for representing various dimensions of a capsule vector. These lanes include different convolution and channel numbers referred to as the depth and width of a lane. This network has faster training and interference and higher accuracy compared to CapsNet.

    \par Xiong et al. \cite{xiong2019deeper} propose Deeper Capsule Network For Complex Data. This network uses Convolutional Capsule Layer in dynamic routing instead of conventional dynamic routing which means replacing the weighted matrix multiplication operation with the convolutional operation. It also has two more convolution Layers for feature extraction and a Caps-Pool as an alternative for max pooling. These methods improve accuracy and optimize the number of parameters. 
    
    \par Phaye et al. \cite{phaye2018dense} propose Dense and diverse capsule networks (DCNet). They employ densely connected convolution layers instead of the standard convolution layers used in Capsnet. This network leads to faster convergence and better performance compared to conventional Capsnet.  

    \par Deliege et al.\cite{deliege2018hitnet} propose HitNet. This network uses a new Hit-or-Miss layer including ghost capsules and prototypes in dynamic routing. In addition, HitNet uses Batch Normalization with an element-wise sigmoid activation function and centripetal loss instead of the squash function and margin loss used in conventional CapsNet. These result in faster and repeatedly better performances compared to CapsNet and an ability to detect potentially mislabeled training images.
    

    \par Our approach stands out from previous works in several key ways. While Deliege et al. \cite{deliege2018hitnet} proposed an enhanced dynamic routing to improve CapsNet, they did not consider parallelism, limiting their ability to fully leverage the potential of parallel processing. Similarly, Rosario et al. \cite{do2019multi} explored parallelism, but not specifically in the context of dynamic routing. In contrast, our approach incorporates both dynamic routing and parallelism in a connected branch manner, which sets it apart from the existing CapsNet variants. This unique combination allows our approach to harness dynamic routing and parallelism benefits, leading to improved performance and energy efficiency.

    \section{Background}
    \par CapsNet replaces scalar-output neurons with groups of neurons referred to as vector-output capsules.
    First, CapsNet applies a two-layer feature extraction to the image to produce representative vectors (capsules). Second, the resulting vectors are encoded in terms of spatial locality through matrix multiplication. This produces Primary Capsules (PCs).


    \par At the next stage, dynamic routing(DR), an alternative to max-pooling, creates output capsules from PCs through a process referred to as routing-by-agreement. DR allocates and manipulates coefficients that define the relation between PCs and output capsules. An iterative process is used in this algorithm to update the coefficients according to agreements among these input and output capsules.
    
    \par These output capsules are used for image reconstruction in a decoder network.
    CapsNet achieves regularization by adding the reconstruction loss (provided by comparing the resulting and the input image) to the loss function.

    
    \par
\begin{figure}[htp]
        \centering
        \includegraphics[keepaspectratio, scale=0.35]{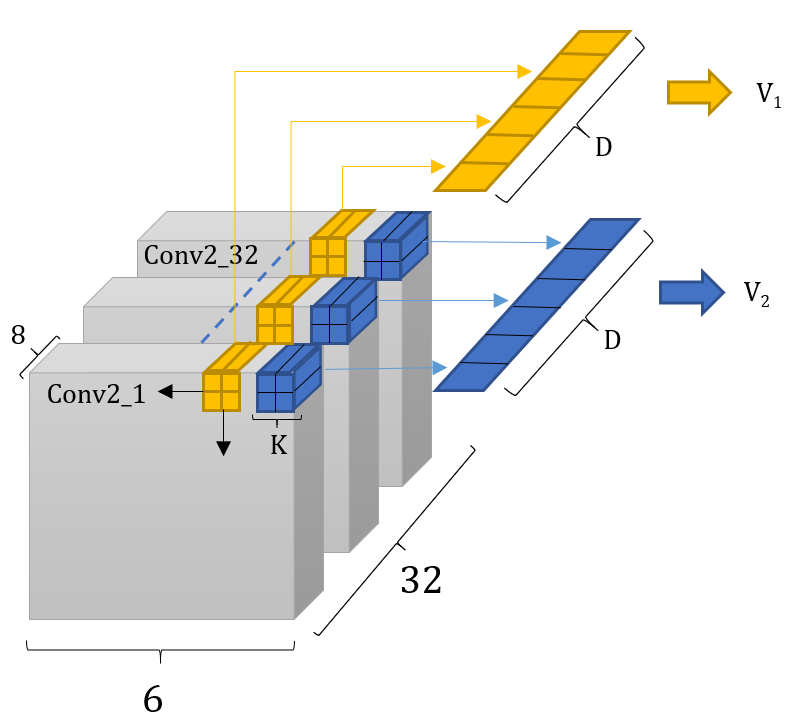}
        \caption{ Convolutional Fully Connected (CFC) layer. In this layer, capsules are created from the translation of neurons. \cite{eshiri2021convolutional} }
        \label{fig:cfc}
    \end{figure}

 Shiri and Baniasadi \cite{eshiri2021convolutional} propose the Convolutional Fully Connected (CFC) layer which we use in PDR-CapsNet. This layer appears after the feature extraction in which capsules are produced (Figure \ref{fig:cfc}). All neurons in different channels, which are known as spatially correlated, are translated to capsule vectors via fully connected layers. They create every capsule from one or multiple neurons in the first channel and their spatially correlated neurons in other channels of the input according to the kernel size selection in CFC.  This preserves the part-to-whole relationship between sub-objects and the main object.

    \section{Method}
    \par  Increasing network complexity can result in overfitting. Our goal is to have high accuracy both on the train and test data. This means that the network must be generalized properly. Regularization is a set of techniques that enhances network generalization and prevents overfitting.
    As shown in Figure \ref{fig7.png}, we used multiple paths to make the network deeper and simultaneously more regularized. In essence, the added paths (represented by side branches) utilize hidden layers and make predictions in parallel. This approach facilitates extracting both low-level and high-level features, improving regularization. 
        

    \par
\begin{figure*}[htp]
        \centering
        \includegraphics[keepaspectratio,scale=0.35]{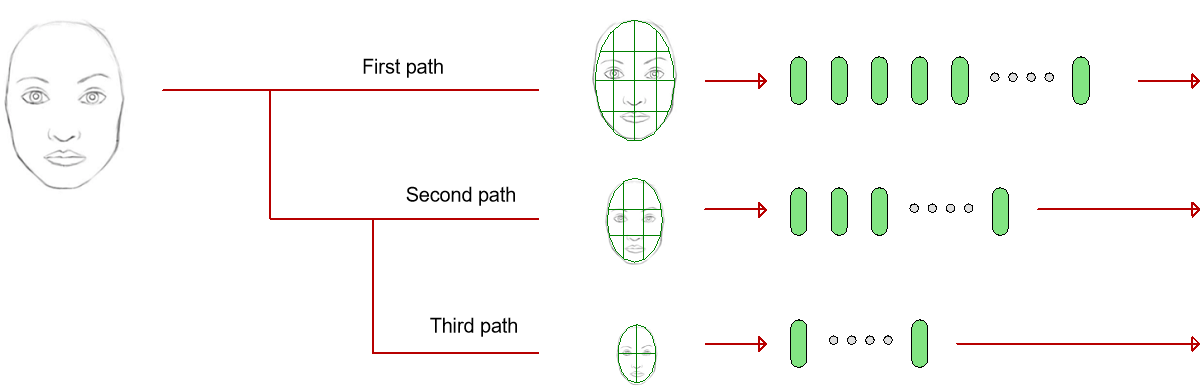}
        \caption{ Capsules with various scales of representation. The third and deepest path results in fewer capsules, each representing larger parts of the image. There is a correlation among path depth, capsule number, and representation scale. }
        \label{fig6.png}
    \end{figure*}

\begin{figure*}[htp]
        \centering
        \includegraphics[keepaspectratio,scale=0.35]{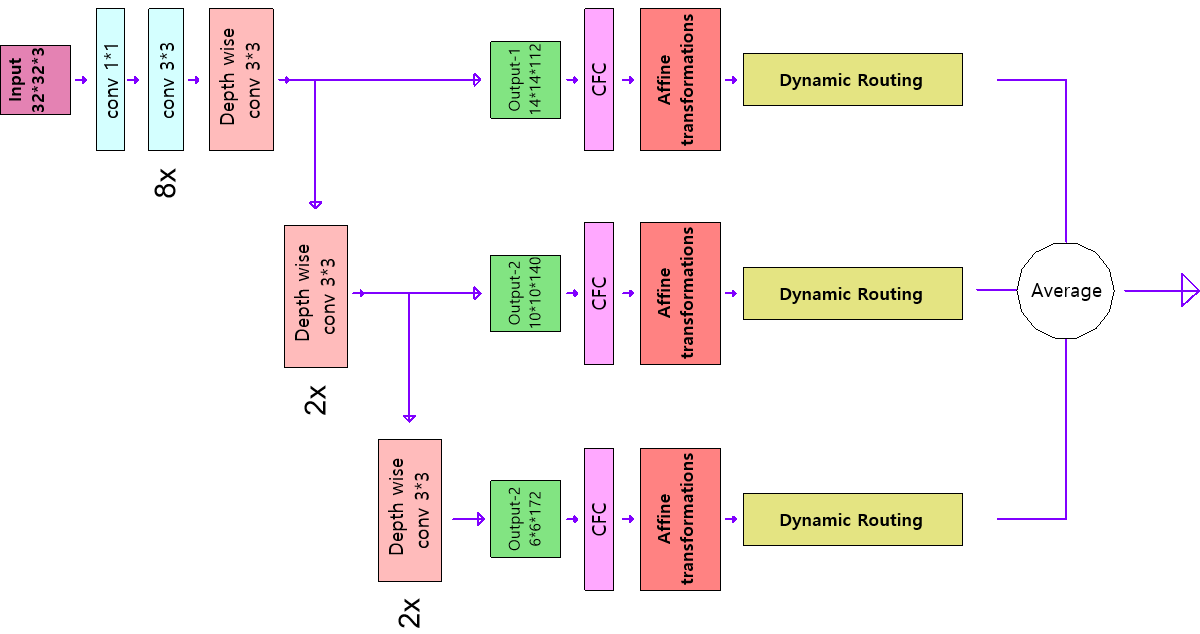}
        \caption{  An overview of the PDR-CapsNet. We use multiple paths with different depths of feature extraction representing small, medium, and large regions of the image. The lowest path is the deepest one as it has two layers more than the first path and one more than the second one. }
        \label{fig7.png}
    \end{figure*}

 Conventional CapsNet utilizes parts to create the whole. In deeper paths, those with more layers, each neuron represents larger regions of the image. This is because deeper paths result in smaller feature maps and fewer neurons representing the input image (Figure \ref{fig6.png}). For example, as Figure \ref{fig7.png} shows, the third path, which is the deepest one, results in a 6*6 feature map, while the first and shallowest path results in a 14*14 feature map. Smaller feature maps result in fewer capsules where each capsule represents larger regions of the image. Capsules representing small, medium, and large regions of the image are fed to three parallel dynamic routing units  which collaborate in creating the whole image. As figure \ref{fig6.png} shows that the first path produces capsules representing smaller parts of the image while the third path produces capsules representing larger regions of the image. Dietterich et al. \cite{dietterich2000ensemble} inspired us to average classification capsules after parallel dynamic routings. Figure \ref{fig7.png} reports the overview of our proposed network.

    \par The original CapsNet uses two 9*9 convolutional layers for feature extraction\cite{sabour2017dynamic}. We improve computational efficiency by replacing 9*9 convolution layers with 3*3 ones. However, utilizing smaller kernel sizes leads to a decline in receptive field\cite{szegedy2016rethinking}. 
    We pay this cost as a price for the additional computations resulting from our more complex network due to using three branches. 
    Additionally, we utilize depth separable convolutions\cite{dietterich2000ensemble} rather than traditional convolutions. We do so primarily to decrease the number of parameters and, secondly, assign different weights to different channels of feature maps in depth-wise convolution. The second approach aims to emphasize different aspects of the image, including both more significant features, which are considered more relevant to the task, and less significant features, which may include background or contextual elements that are not as critical (Figure \ref{fig: depth}).
    

    \par Training is usually challenging as the parameters of previous layers change the input distribution of later layers. This change is referred to as Internal Covariate Shift\cite{ioffe2015batch} and occurs in CapsNets too. By adopting Batch Normalization after convolutions, we fix the distribution of the layers' inputs to reduce the Internal Covariate Shift and accelerate training.

    \section{Configurations}
    \subsection{Datasets}
     CapsNet and its variants do not usually perform well for large-scale datasets. We test PDR-CapsNet on small-scale datasets, i.e., Fashion-MNIST (F-MNIST) \cite{xiao2017fashion}, SVHN \cite{netzer2011reading} and CIFAR-10 \cite{krizhevsky2009cifar} datasets. Table \ref{table: table_dsets} shows the properties of these datasets. F-MNIST consists of 28x28 grayscale images of 60,000 fashion products from 10 categories, and the training and testing sets include 50,000 and 10,000 images, respectively. SVHN and CIFAR-10 both have 32x32 RGB images from 10 categories. Training and testing test sizes of SVHN are 73,257 and 26,032, while those of CIFAR-10 are 50,000 and 10,000, respectively. Cropped digits of the house numbers images are included in the SVHN dataset.

    \begin{table}[htp]
    \caption{Datasets used to test CFC-CapsNet.}
    \centering
    \resizebox{\linewidth}{!} {\begin{tabular}[width=\linewidth]{|c | c | c | c | c|} 
    \hline\hline
    \textbf{Name} & \textbf{Image Size} & \textbf{\#Channels} & \textbf{Training samples} & \textbf{Test Samples}  \\ [1ex]
    \hline
    F-MNIST  & 28x28  & 1  & 50,000 & 10,000 \\ [1ex]
    \hline
    SVHN & 32x32 & 3 & 73,257 & 26,032 \\[1ex]
    \hline
    CIFAR-10 & 32x32 & 3 & 50,000 & 10,000 \\[1ex] 
    \hline
    \end{tabular}}
    \label{table: table_dsets}
    \end{table}

\subsection{Experiments settings} 
    We implement PDR-CapsNet on top of the PyTorch implementation of CapsNet \footnote{https://github.com/gram-ai/capsule-networks}. We use a 2080Ti GPU with 11GB VRAM to run the experiments. 
    We train the network twice through a method referred to as hard training. In the second round of this method, we use tighter bonds in the loss function. We repeat the experiments five times and report the average values as we observe a slight variation in results. We use the Adam optimizer and the default learning rate of LR=0.001. Moreover, we use an exponential decay of gamma=0.96 for the learning rate. The batch size is set to 128.

    \subsection{Accuracy of the Network}  
    Table \ref{table: accuracy} reports PDR-CapsNet's accuracy. As this table demonstrates, PDR-CapsNet achieves 11.6\%, 1.93\%, and 1.39\% higher accuracy for CIFAR-10, SVHN, and F-MNIST datasets respectively, and compared to CapsNet. This improvement is due to the enhanced capsules produced through branches with different depths of feature extraction. These branches overcome underfitting by providing the network with various paths. As shown in Table \ref{table: accuracy}, PDR-CapsNet is more accurate compared to some of the state-of-the-art CapsNets.
    
    \begin{table*}[htp]
    \caption{Comparison of accuracy in various CapsNets.}
    \centering
    \resizebox{\linewidth}{!}{\begin{tabular}[width=\linewidth]{|c | c | c | c |} 
    \hline\hline
    \textbf{Architecture } & \textbf{Accuracy (CIFAR-10)} & \textbf{Accuracy (SVHN)} & \textbf{Accuracy (F-MNIST)} \\[1ex]
    \hline
    CapsNet(Pytorch) \cite{sabour2017dynamic}  & 71.69\%  & 92.70\% & 91.37\% \\ [1ex]
    \hline
    CFC-CapsNet\cite{eshiri2021convolutional}  & 73.85\% & 93.30\% & 92.86\% \\[1ex]
    \hline
    MS-CapsNet \cite{xiang2018ms}  & 72.30\% & 92.68\% & 92.26\% \\[1ex] 
    \hline
    MLCN \cite{do2019multi} & 75.18\% & - & 92.63\% \\[1ex] 
    \hline
    LE-CapsNet \cite{shiri2021capsnet}  & 75.75\% & 93.02\% & 93.14\% \\[1ex] 
    \hline
    LE-CapsNet(+ Dropout) \cite{shiri2021capsnet}  & 76.73\% & 92.62\% & 93.04\% \\[1ex] 
    \hline
    DeeperCaps\cite{xiong2019deeper}   & 81.29\% & - & - \\[1ex] 
    \hline
    DCNet \cite{phaye2018dense}  & 82.63\% & 95.58\% & 94.64\% \\[1ex] 
    \hline
     DCNet++ \cite{phaye2018dense}  & 89.71\% & 96.90\% & 94.65\% \\[1ex] 
    \hline
    HitNet\cite{deliege2018hitnet}   & 73.30\% & 94.5\% & 92.3\% \\[1ex] 
    \hline
    PDR-CapsNet & \textbf{83.55} \% & \textbf{94.63}\% & \textbf{92.76}\% \\[1ex] 
    \hline
    \end{tabular}}
    \label{table: accuracy}
    \end{table*}

    \subsection{The Confusion matrix of the network}  
\begin{figure*}[htp]
        \centering
        \includegraphics[keepaspectratio,scale=0.63]{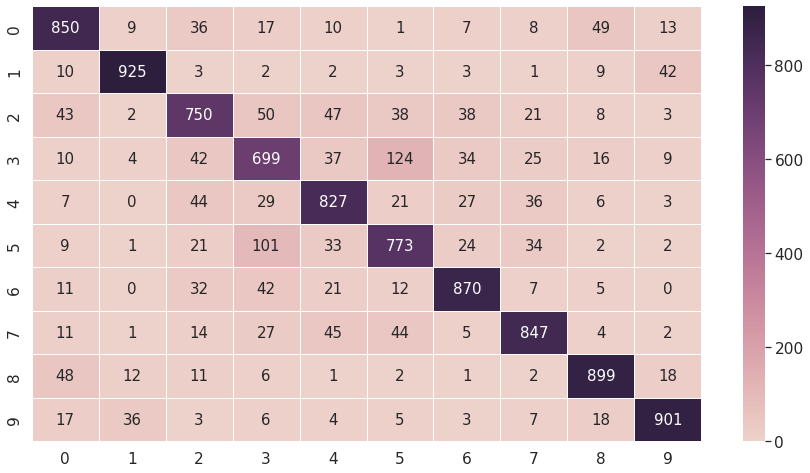}
        \caption{ The confusion matrix of the PDR-CapsNet on the CIFAR-10 dataset.}
        \label{fig: confmat_PDR}
        \end{figure*}

\begin{figure*}[htp]
        \centering
        \includegraphics[keepaspectratio,scale=0.63]{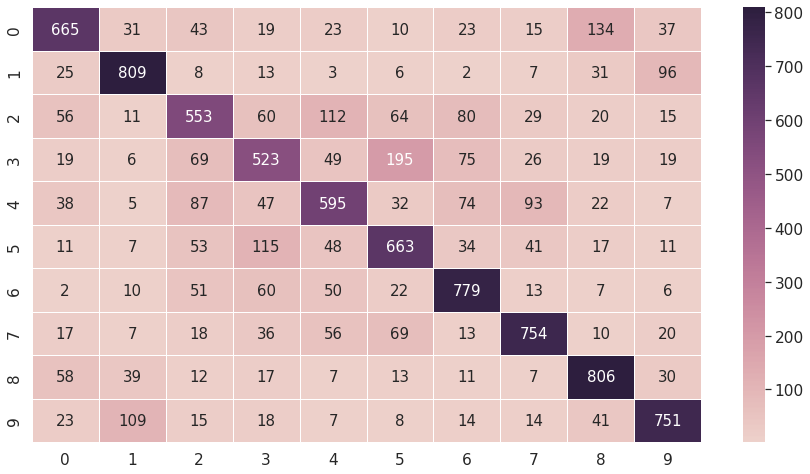}
        \caption{ The confusion matrix of the original CapsNet on the CIFAR-10 dataset.  }
        \label{fig: confmat-capsnet}
        \end{figure*}

    The Confusion matrix assists us in interpreting how our model wrongly classifies each of the categories. The visualization of the normalized Confusion matrices of our model and the CapsNet on the CIFAR-10 dataset are presented in Figures \ref{fig: confmat_PDR} and \ref{fig: confmat-capsnet}, respectively. In the provided confusion maps, (0, 1, 2, 3, 4, 5, 6, 7, 8, 9) are the class labels presenting (airplane, automobile, bird, cat, deer, dog, frog, horse, ship, truck), respectively. The horizontal axis corresponds to the actual labels in these maps, while the vertical axis represents the predicted classes.
    \par As Figure \ref{fig: confmat_PDR} shows, 7 classes achieved accuracies higher than 82.5\%. We notice that our network sometimes confuses the cat class with the dog one, making the accuracy of the cat class 69.9\% which is the lowest. This could be due to the fact that the two classes have comparable features, leading to the network making similar classifications.

     \par For the original CapsNet, 5 classes achieved accuracies greater than 75\%, in which the higher achieved accuracy is 80.9\%. Furthermore, the model confuses the cat and the dog class 19.5\%.

    \subsection{ Precision, Recall and F1-score }
    We use Precision, Recall and F1-score as classification metrics. As such, we evaluate the extent to which our network is able to distinguish positive samples from negative ones and detect all positive samples. PDR-CapsNet outperforms CapsNet by approximately 14\% in terms of precision, recall, and F1-score Table \ref{table: Precision, Recall and F1-score}.

     \begin{table}[htp]
    \caption{Comparison of Precision, Recall and F1-score of the original CapsNet and PDR-CapsNet on the CIFAR-10 dataset.}
    \centering
    \small
    \resizebox{\linewidth}{!}{\begin{tabular}[width=\linewidth]{|c | c | c | c |} 
    \hline\hline
    \textbf{Architecture } & \textbf{Precision \%\ } & \textbf{Recall \%\ } & \textbf{F1-score \%\ }\\[1.5ex]
    \hline
    CapsNet   & 68.8 & 69  & 68.8 \\ [1.5ex]
    \hline
    
    PDR-CapsNet &  \textbf{83.4} &  \textbf{83.4} &  \textbf{83.4}\\[1.5ex] 
    \hline
    \end{tabular}}
    \label{table: Precision, Recall and F1-score}
    \end{table}

    \subsection{Number of Parameters}

   Table \ref{table: parameter} reports the number of parameters. PDR-CapsNet requires 87.26 \% fewer parameters compared to CFC-CapsNet for CIFAR-10. It also includes fewer computations compared to CapsNet. This reduction is mostly due to employing the depth-wise separable convolutions in later layers and 3*3 convolutions in primary layers.
    
     \begin{table}[htp]
    \caption{Comparison of the number of parameters in various CapsNets.}
    \centering
    \small
    \begin{tabular}{|c | c |} 
    \hline\hline
    \textbf{Architecture } & \textbf{Parameter Number (CIFAR-10)}  \\[0.5ex]
    \hline
    CapsNet   & 11.7M  \\ [0.5ex]
    \hline
    CFC-CapsNet  & 5.9M  \\[0.5ex]
    \hline
    MS-CapsNet   & 13.9M  \\[0.5ex] 
    \hline
    MLCN   & 14.2M  \\[0.5ex] 
    \hline
    LE-CapsNet   & 3.8M  \\[0.5ex] 
    \hline
    DeeperCaps   & 5.81M  \\[0.5ex] 
    \hline
    DCNet   & 11.88M  \\[0.5ex] 
    \hline
    HitNet   & 8.89M  \\[0.5ex] 
    \hline
    PDR-CapsNet &  \textbf{1.49M}\\[0.5ex] 
    \hline
    \end{tabular}
    \label{table: parameter}
    \end{table}

    \subsection{Number of MACs and Flops} 
    \label{sec: MACs and FLOPs}


    Table \ref{table: MACs} compares PDR-CapsNet's number of MACs and FLOPs to CapsNet. As this Table demonstrates, PDR-CapsNet requires 32.27\%, 47.40\%, fewer MACs, and Flops for CIFAR-10 compared to CapsNet (the batch size of both networks is 128 in this experiment).
    
    
    This significant improvement can be attributed to the reduced number of capsules and the enhanced feature extraction structure employed by PDR-CapsNet. These changes make the network less computationally expensive while improving its accuracy.
    
    Note that the number of MACs and FLOPs can serve as an indirect and
    implementation-independent measurement of energy consumption.

    \begin{table}[htp]
    \caption{Comparison of the number of MACs and FLOPs in original CapsNet and PDR-CapsNet.}
    \centering
    \small
    \resizebox{\linewidth}{!}{\begin{tabular}[width=\linewidth]{|c | c | c |} 
    \hline\hline
    \textbf{Architecture } & \textbf{MACs Number (CIFAR-10)} & \textbf{FLOPs Number (CIFAR-10)} \\[1.5ex]
    \hline
    CapsNet   & 48.25G & 96.67G  \\ [1.5ex]
    \hline
    
    PDR-CapsNet &  \textbf{32.68G} &  \textbf{65.58G}\\[1.5ex] 
    \hline
    \end{tabular}}
    \label{table: MACs}
    \end{table}

    \subsection{Network Training and Testing Time } 
\begin{figure*}[htp]
        \centering
        \includegraphics[keepaspectratio,scale=0.5]{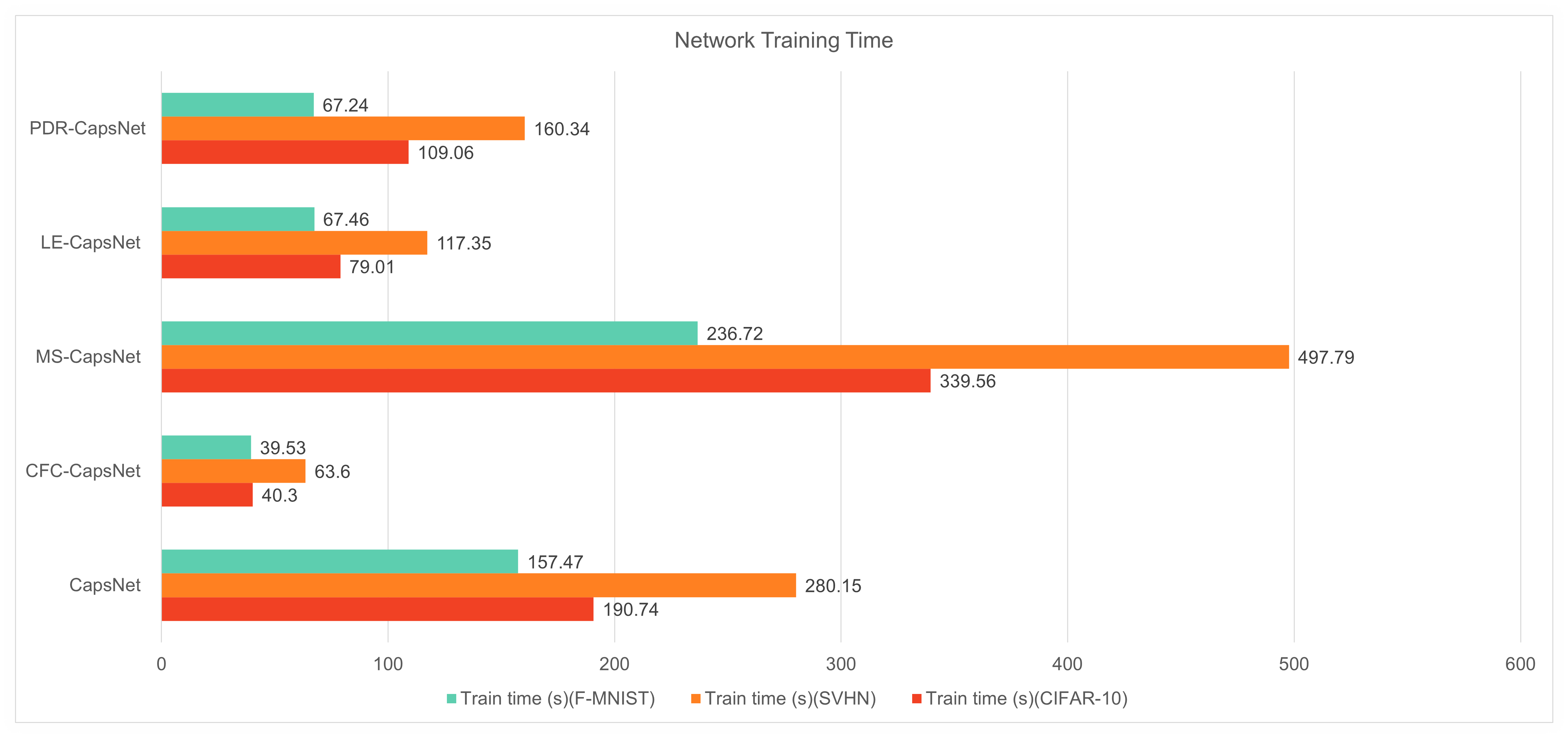} 
        \caption{  Network training time for different CapsNet variations. PDR-CapsNet obtains competitive results.
    }
    \label{fig: train time}
    \end{figure*}

\begin{figure*}[htp]
        \centering
        \includegraphics[keepaspectratio,scale=0.5]{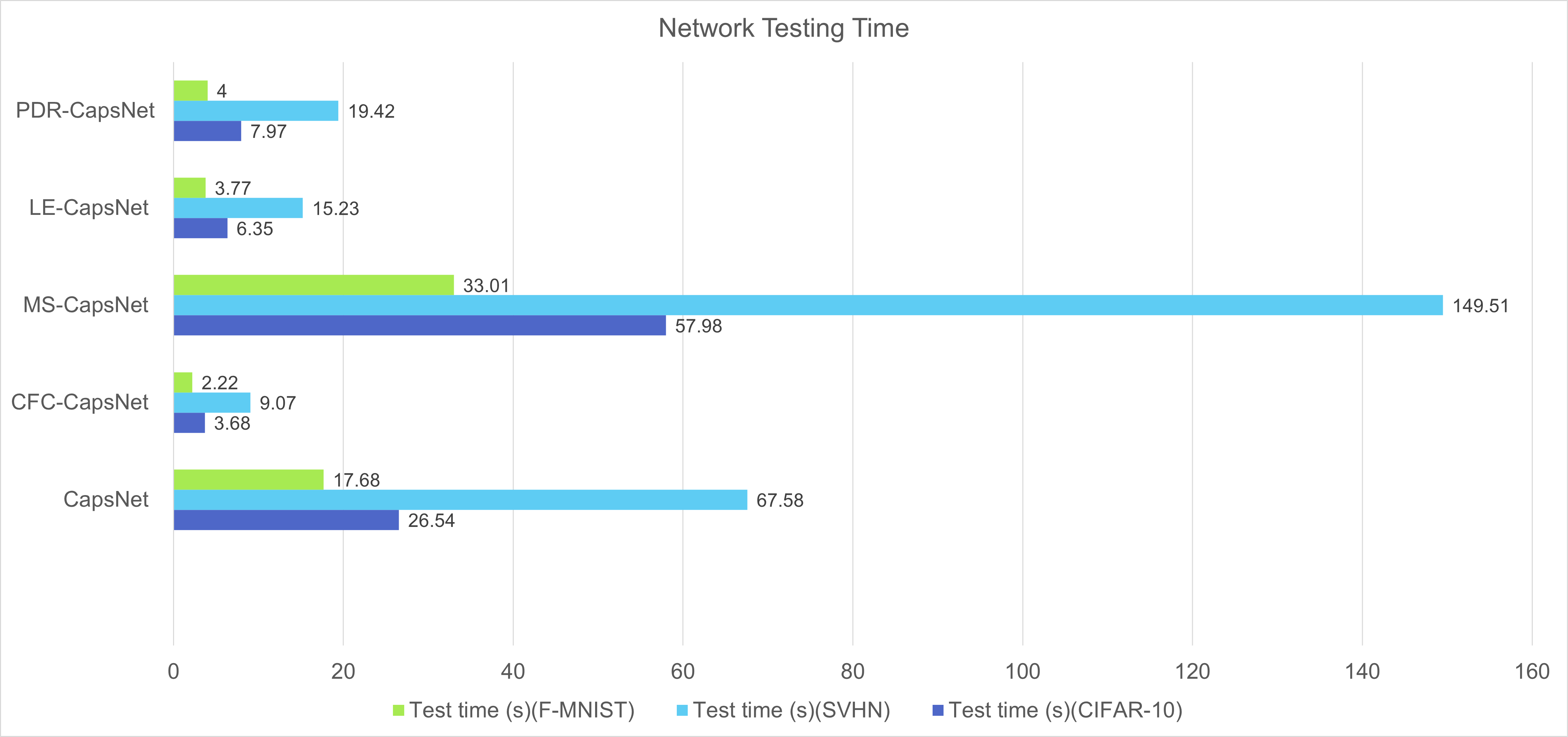}
        \caption{  Network testing time for different CapsNet variations. PDR-CapsNet achieves competitive results.
    }
    \label{fig: test time}
    \end{figure*}

   Figures \ref{fig: train time} and \ref{fig: test time} show some state-of-the-art CapsNets' training and testing times. In addition, we report the number of capsules in Table \ref{table: the number of capsules}. PDR-CapsNet utilizes fewer capsules compared to CapsNet, resulting in a less computationally expensive process, particularly in dynamic routing. This is why PDR-Capsnet is faster compared to  CapsNet(1.75x and 3x faster training and testing respectively). However, with 332 capsules for CIFAR-10 and SVHN datasets and 140 for the F-MNIST dataset, PDR-CapsNet employs more capsules compared to CFC-CapsNet and LE-CapsNet due to its different feature extraction structure. Thus, PDR-CapsNet is not as fast as CFC-CapsNet and LE-CapsNet.

 \begin{table}[htp]
    \caption{The number of capsules.\\ Different feature extraction structures and input image sizes impact the number of primary capsules }
    \centering
    \resizebox{\linewidth}{!}{\begin{tabular}[width=\linewidth]{|c | c | c | c |} 
    \hline\hline
    \textbf{Architecture } & \textbf{CIFAR-10} & \textbf{SVHN} & \textbf{F-MNIST} \\[1ex]
    \hline
    CapsNet \cite{sabour2017dynamic}  & 2048 & 2048 & 1152 \\ [1ex]
    \hline
    CFC-CapsNet\cite{eshiri2021convolutional}  & 64 & 64 & 36 \\[1ex]
    \hline
    MS-CapsNet \cite{xiang2018ms}  & 757 & 757 & 513 \\[1ex] 
    \hline
    LE-CapsNet \cite{shiri2021capsnet}  & 192 & 192 & 108 \\[1ex] 
    \hline
    PDR-CapsNet & \textbf{332}  & \textbf{332} & \textbf{140} \\[1ex] 
    \hline
    \end{tabular}}
    \label{table: the number of capsules}
    \end{table}

  \subsection{Energy Consumption } 

\begin{figure*}[htp]
        \centering
        \includegraphics[keepaspectratio,scale=0.5]{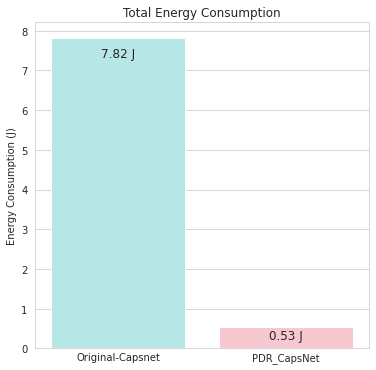}
        \caption{  Comparison of total energy consumption for CapsNet and PDR-CapsNet.
    }
    \label{fig: enery_consumption}
    \end{figure*}

    Figure \ref{fig: enery_consumption} compares the total energy consumption of PDR-CapsNet to CapsNet. PDR-CapsNet consumes 0.53 J of energy, while CapsNet consumes 7.82 J of energy. The energy consumption data for these networks were obtained by monitoring the GPU power usage using a GPU monitoring tool. The power usage data was then used to calculate the energy consumption of the networks over the entire duration of execution. We used early stopping during training to prevent overfitting and achieve faster convergence. We stop both PDR-CapsNet and CapsNet after converging to a similar accuracy level (CapsNet's accuracy). PDR-CapsNet requires less training and inference time and has fewer MACs and FLOPs, resulting in lower energy consumption compared to CapsNet. This is consistent with the results reported earlier in section \ref{sec: MACs and FLOPs} where PDR-CapsNet produced less number of MACs and FLOPs.

   \subsection{Network Complexity } 
   We introduce a new metric, Network Complexity (NC) to produce better insight. Test time and accuracy are influenced by both the number of capsules and computational efficiency. NC can be considered as a metric providing a deeper understanding of this trade-off.  We define NC as follows:


     \begin{equation}
      NC=  Test \;time * (Error-rate ^ \frac{1}{n}) )\label{eq1}
     \end{equation}

     The test time is reported in Figure \ref{fig: test time}.  

     \begin{equation} 
     Error rate = (1-accuracy)\label{eq2}
     \end{equation}
     
     The accuracy is reported in Table \ref{table: accuracy}.
     
     n is a parameter representing different importance degrees of accuracy or test time in CapsNets. \( n<1\)  means the network places more importance on accuracy compared to test time. However, \( n>1\)  shows that networks pay more attention to test time reduction. \\
     
\begin{figure}[htp]
        \centering
        \includegraphics[keepaspectratio,scale=0.45]{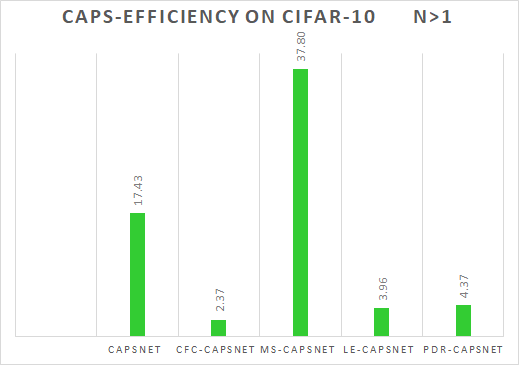}
        \caption{  Network Complexity of various CapsNets when \(n>1\).
        Our network shows less NC compared to CapsNet and MS-CapsNet.
    }
    \label{fig:n>1}
    \end{figure}

\begin{figure}[htp]
        \centering
        \includegraphics[keepaspectratio,scale=0.45]{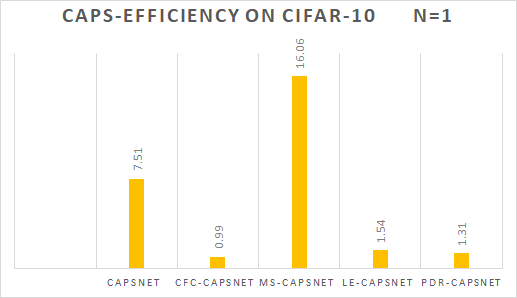}
        \caption{  Network Complexity when n=1.
        Our network shows less NC compared to CapsNet, MS-CapsNet and LE-CapsNet.
    }
    \label{fig:n=1}
    \end{figure}

\begin{figure}[htp]
        \centering
        \includegraphics[keepaspectratio,scale=0.45]{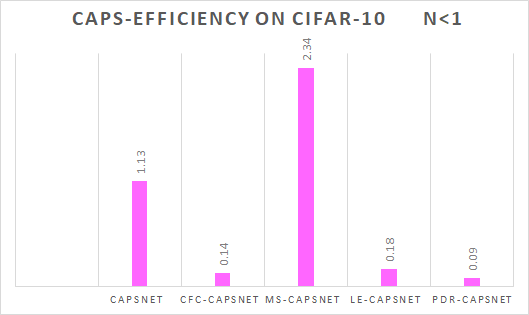}
        \caption{  Network Complexity of various CapsNets when \(n<1\).
        Our network shows lower NC compared to CapsNet, MS-CapsNet, LE-CapsNet and CFC-CapsNet.
    }
    \label{fig:n<1}
    \end{figure}

     Less NC indicates better network performance. Figures \ref{fig:n>1},\ref{fig:n=1} and \ref{fig:n<1} report NC for various CapsNets with different n values. The figures show that CapsNet has the highest NC regardless of the n value. Figure \ref{fig:n>1} shows that when we choose \(n>1\) CFC-CapsNet and LE-CapsNet outperform PDR-CapsNet. 
     However, for n=1, Figure \ref{fig:n=1}, PDR-CapsNet outperforms LE-CapsNet. In addition, for \(n<1\), PDR-CapsNet does better than CFC-CapsNet and LE-CapsNet. Accordingly, PDR-CapsNet focuses more on accuracy reduction compared to test time reduction.

    \subsection{Convergence} 
\begin{figure}[ht]
        \centering
        \includegraphics[keepaspectratio,scale=0.5]{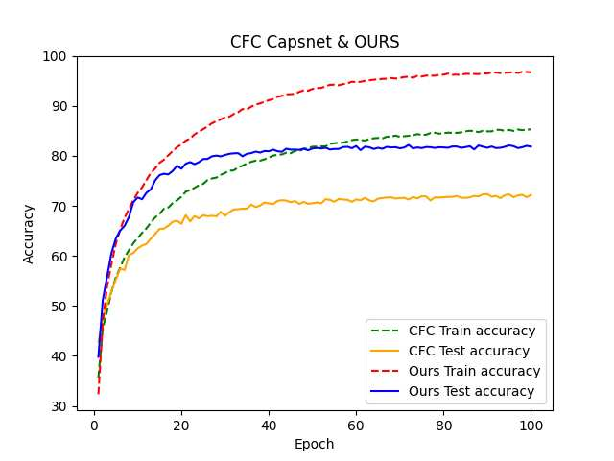}
        \caption{ CFC-Capsnet and PDR-CapsNet train and test accuracy for the CIFAR-10 dataset.}
    \label{fig: Convergence}
    \end{figure}

    Figure \ref{fig: Convergence} reports the result of  PDR-CapsNet's test and train accuracy compared to CFC-Capsnet for the CIFAR-10 dataset. As reported, PDR-CapsNet performs better in training and avoiding overfitting compared to CFC-Capsnet.
    
\section{CONCLUSION}
\par 

    We propose  Parallel Dynamic Routing CapsNet and improve energy efficiency while reducing underfitting, and maintaining the generalization of CapsNet. By introducing branches to produce enhanced capsules, we improved performance and also added regularization to the network. We used depth-wise separable and 3*3 convolutions to reduce computational complexity and achieve faster inference. PDR-CapsNet achieved a higher accuracy of 83.55\% for CIFAR-10 while consuming significantly lower energy compared to CapsNet. Specifically, PDR-CapsNet consumed only 0.53J of energy (vs. 7.82J consumed by CapsNet) achieving competitive accuracy( 71.69\% for CIFAR-10). Furthermore, PDR-CapsNet employs fewer capsules and parameters, making it a more efficient and practical solution for deep learning models.

    \section*{Acknowledgment}
    This research has been funded in part or completely by the Computing Hardware for Emerging Intelligent Sensory Applications (COHESA) project. COHESA is financed under the National Sciences and Engineering Research Council of Canada (NSERC) Strategic Networks grant number NETGP485577-15.


\bibliographystyle{IEEEtranS}
\bibliography{refs}

\end{document}